\documentclass{article}
\usepackage{arxiv}

\usepackage[utf8]{inputenc} 
\usepackage[T1]{fontenc}    
\usepackage{hyperref}       
\usepackage{url}            
\usepackage{booktabs}       
\usepackage{amsfonts}       
\usepackage{nicefrac}       
\usepackage{microtype}      
\usepackage{lipsum}
\usepackage{subcaption}

\usepackage{relsize}
\usepackage{verbatim}
\usepackage{graphicx}
\usepackage{dcolumn}
\usepackage{bm}
\usepackage{float}

\usepackage{graphics}
\usepackage{color}
\usepackage{xcolor}
\usepackage{bm}

\usepackage{soul}

\usepackage{amsmath,amssymb,amsfonts}

\usepackage{algorithm}
\usepackage{algpseudocode}
\algnewcommand\server{\item[\textbf{Server execution:}]}%
\algnewcommand\client{\item[\textbf{ClientUpdate($k,w$):}]}%

\usepackage{enumitem}
\setlist[itemize]{leftmargin=*}
\setlist[enumerate]{leftmargin=*}
\usepackage{amsmath,amssymb,amsfonts}
\usepackage{graphics}
\usepackage{color}
\usepackage{xcolor}
\definecolor{rev}{rgb}{0,0,0}
\definecolor{rev2}{rgb}{0,0,0}

\usepackage{array}
\newcolumntype{P}[1]{>{\centering\arraybackslash}p{#1}}
\usepackage{multirow}
\usepackage{cancel}
\usepackage[capitalise]{cleveref} 
\usepackage{enumitem}

\usepackage{cite}

\usepackage{amsmath, amssymb, graphicx, subcaption, booktabs, multirow, natbib}
\usepackage{adjustbox}
\usepackage{xcolor}
\title{Localized PCA-Net Neural Operators for Scalable Solution Reconstruction of Elliptic PDEs}

\author{
  Mrigank Dhingra \\
  Department of Mechanical and Aerospace Engineering\\
  University of Tennessee, Knoxville\\
  Knoxville, TN 37996, USA.\\
  \texttt{mdhingra@vols.utk.edu}\\
  \And
  Romit Maulik \\
  Informatics and Intelligent Systems\\
  The Pennsylvania State University\\
  University Park, PA 16802, USA.\\
  \texttt{rmaulik@psu.edu}\\
  \And
  Adil Rasheed\\
  Department of Engineering Cybernetics\\
  Norwegian University of Science and Technology\\
  Trondheim, NO-7465 Norway. \\
  \texttt{adil.rasheed@ntnu.no}\\
  \And
  Omer San \\
  Department of Mechanical and Aerospace Engineering\\
  University of Tennessee, Knoxville\\
  Knoxville, TN 37996, USA.\\
  \texttt{osan@utk.edu} \\
}

\date{}

\definecolor{ProfSanColor}{RGB}{0,0,255}     
\definecolor{MrigankColor}{RGB}{255,0,0}     
\definecolor{ArthColor}{RGB}{0,128,0}        
\definecolor{BipinColor}{RGB}{128,0,128}     
\definecolor{AbidColor}{RGB}{255,165,0}      
\definecolor{AnilColor}{RGB}{255,105,180}    

\begin{document}

\maketitle

\begin{abstract}

Neural operator learning has emerged as a powerful approach for solving partial differential equations (PDEs) in a data-driven manner. However, applying principal component analysis (PCA) to high-dimensional solution fields incurs significant computational overhead. To address this, we propose a patch-based PCA-Net framework that decomposes the solution fields into smaller patches, applies PCA within each patch, and trains a neural operator in the reduced PCA space. We investigate two different patch-based approaches that balance computational efficiency and reconstruction accuracy: (1) local-to-global patch PCA, and (2) local-to-local patch PCA. The trade-off between computational cost and accuracy is analyzed, highlighting the advantages and limitations of each approach. Furthermore, within each approach, we explore two refinements for the most computationally efficient method: (i) introducing overlapping patches with a smoothing filter and (ii) employing a two-step process with a convolutional neural network (CNN) for refinement. Our results demonstrate that patch-based PCA significantly reduces computational complexity while maintaining high accuracy, reducing end-to-end pipeline processing time by a factor of 3.7 to 4× compared to global PCA, thefore making it a promising technique for efficient operator learning in PDE-based systems.
\end{abstract}

\section{Introduction}

Numerical simulation and analysis of PDEs play a central role across numerous fields, including fluid dynamics, heat transfer, and electromagnetics. Traditional PDE solvers, such as finite-difference and finite-element methods, often require significant computational resources, especially for high-dimensional systems or large-scale domains \cite{brandt1977multi, blanusa2022roleinternalvariabilityglobal, quarteroni2008numerical}. Recently, data-driven neural operators have emerged as powerful alternatives, capable of rapidly approximating PDE solutions by learning mappings directly from input conditions to PDE solution spaces \cite{li2021fourierneuraloperatorparametric, lu2021deeponet, kovachki2023neural, HUA202321, kadeethum2023latentNO, TRIPURA2023115783, 10.5555/3648699.3649087, 10.5555/3618408.3618917, 10.1093/nsr/nwad336, RAMEZANKHANI2025109886, GRADY2023105402, seo2022pretraining}.

 PCA is a widely-used dimensionality reduction technique frequently integrated into neural operators to reduce computational complexity and enhance generalization capabilities \cite{SMAI-JCM_2021__7__121_0, 10.5555/3648699.3649017}. However, applying PCA globally across high-dimensional solution fields poses a significant computational bottleneck, primarily due to the expensive singular value decomposition (SVD) step involved. Specifically, performing PCA via SVD on an $m \times n$ data matrix ($m$ samples, each of dimension $n$) entails computational complexity of:
\begin{equation}
\mathcal{O}(\min(mn^2, m^2n)).
\end{equation}

Recent advances in operator learning have increasingly adopted domain decomposition strategies to address the scalability limitations of global neural solvers \cite{klawonn2024mlddm}. These approaches divide the computational domain into smaller, localized regions and train separate neural networks or operators within each subdomain \cite{10.5555/3618408.3618917, LI2025117732}. For example, Finite Basis PINNs (FBPINNs) \cite{Moseley2021FBPINN} decompose the domain into overlapping patches and assign a dedicated sub-network to each, significantly improving convergence by reducing spectral bias and enabling localized feature learning. Extensions such as multilevel FBPINNs \cite{DOLEAN2024117116} and multifidelity PINN decomposition frameworks \cite{heinlein2024multifidelitydomaindecompositionbasedphysicsinformed} further integrate coarse global components and heterogeneous fidelity levels to ensure continuity and scalability across subdomains. Similarly, Schwarz-type methods have been employed to couple multiple physics-informed neural networks (PINNs) across overlapping or non-overlapping subdomains, often learning interface conditions such as Robin boundary terms to enhance communication between patches \cite{feeney2023breakingboundariesdistributeddomain, basir2023generalizedschwarztypenonoverlappingdomain}. In the context of neural operators, localized Fourier Neural Operators (Local-FNO) \cite{QIN2025112668} have demonstrated improved accuracy and efficiency by training operators on overlapping spatial tiles, with additional blending mechanisms to enforce global coherence. These works collectively highlight the benefits of localizing the learning process — enabling reduced memory footprints, distributed training, and more accurate resolution of multiscale features. Motivated by this trend, our proposed framework introduces a patch-based neural operator architecture that operates within a PCA-reduced subspace for each patch. Unlike previous domain decomposition methods which typically learn in the full solution space, we exploit PCA to achieve dimensionality reduction and computational savings within each subdomain. Furthermore, we introduce refinements such as overlapping patch blending and CNN-based post-processing to mitigate reconstruction artifacts and preserve global consistency, thereby bridging the efficiency of localized learning with the fidelity of global reconstructions \cite{6795724, NIPS2012_c399862d}. Tensor-compressed operator learning architectures have provided compelling alternatives to traditional global solvers for high-resolution PDEs. In particular, the Multi-Grid Tensorized Fourier Neural Operator (MG-TFNO) \cite{kossaifi2023multigridtensorizedfourierneural, tran2023factorizedfourierneuraloperators, rahman2023unoushapedneuraloperators, Panagakis_2021, doi:10.1137/07070111X, guibas2022adaptivefourierneuraloperators, li2023physicsinformedneuraloperatorlearning, 10.1145/3592979.3593412} addresses two critical limitations in operator learning: memory overhead and limited generalization in low-data regimes. MG-TFNO achieves significant compression by (i) tensorizing the convolutional weights in the spectral domain using low-rank decompositions such as Tucker and CP, and (ii) decomposing the spatial domain into multigrid-structured patches to enable localized learning and parallelization. These enhancements allow MG-TFNO to outperform standard FNOs and other baselines with over 150$\times$ parameter compression and reduced input dimensionality, all while maintaining or improving prediction accuracy on PDE benchmarks like Navier-Stokes and Burgers' equation. In this work, we directly compare our proposed patch-based PCA-Net neural operator with MG-TFNO. While MG-TFNO reduces dimensionality through spectral-domain tensor factorization, our approach achieves similar compression by leveraging spatial-domain PCA within localized patches. This comparison aims to highlight the trade-offs between spectral versus spatial dimensionality reduction, particularly in terms of computational efficiency, representation quality, and reconstruction smoothness.

\begin{figure}
    \centering
    \begin{subfigure}{0.5\textwidth}
        \centering
        \includegraphics[width=\textwidth]{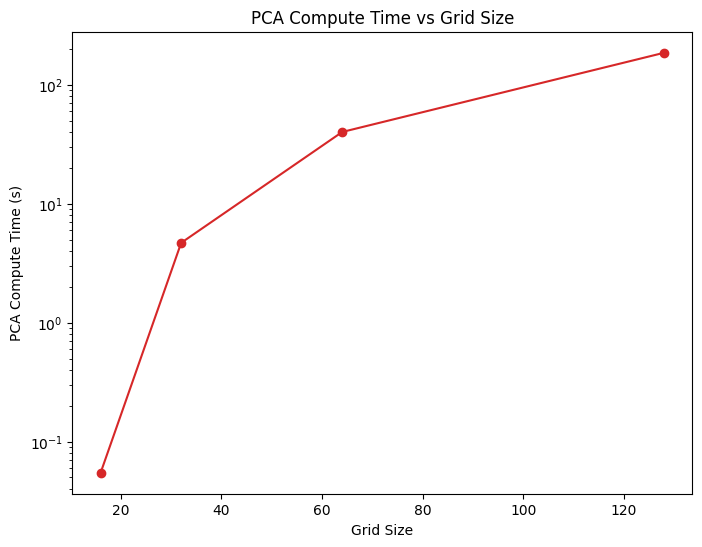}
        \caption{PCA Compute Time vs Grid Size}
    \end{subfigure}


    \begin{subfigure}{0.5\textwidth}
        \centering
        \includegraphics[width=\textwidth]{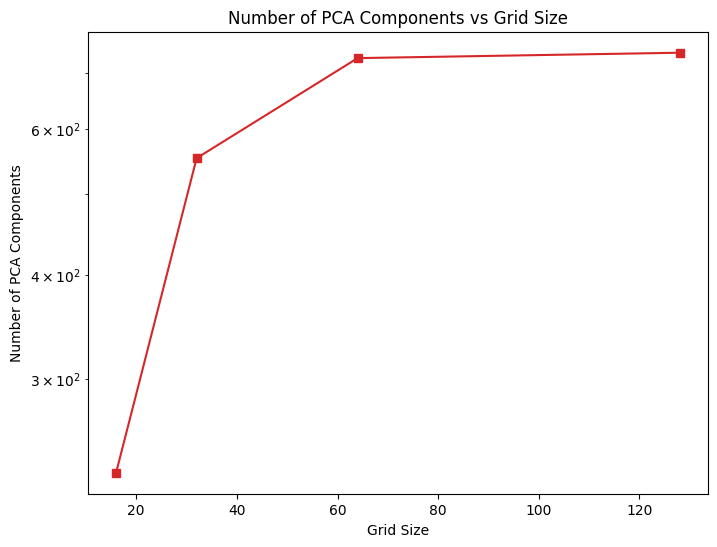}
        \caption{Number of PCA Components vs Grid Size}
    \end{subfigure}



    \caption{Computational Costs of PCA on Different Grid Sizes. Each plot highlights a separate metric: compute time, and number of PCA components.}
    \label{fig:pca_comparison}
\end{figure}

For PDE solutions discretized on fine grids (e.g., $128 \times 128$), the dimension $n$ is large, leading to prohibitively high computational costs. This complexity becomes especially critical when processing large datasets containing thousands of samples, making the global PCA approach impractical for scalable neural operator frameworks (Fig. \ref{fig:pca_comparison} shows the increase in compute times with the domain size).

To address these challenges, we propose a patch-based PCA-Net neural operator framework, wherein the solution domain is partitioned into smaller, manageable patches \cite{XIAO201915, WANG2022114424}. PCA is subsequently performed independently on each patch. By decomposing the original large data matrix into smaller patch matrices, the computational complexity reduces substantially. If we partition each solution field of dimension $D \times D$ into patches of dimension $p \times p$, we obtain $\frac{D^2}{p^2}$ patches per sample. Conducting PCA on each patch independently yields complexity per patch as:
\begin{equation}
\mathcal{O}\left(\frac{D^2}{p^2} \cdot \min(mp^4, m^2p^2)\right).
\end{equation}
Since typically $p \ll D$, this significantly reduces total computational overhead compared to global PCA.

While this patch-based approach efficiently addresses the computational bottleneck, independent processing of patches may introduce reconstruction artifacts, primarily manifesting as discontinuities or blockiness at patch boundaries. To mitigate this, we explore two key refinements: (i) introducing overlapping patches coupled with smoothing filters (e.g., the Hanning window) to blend reconstructions smoothly \cite{2020SciPy-NMeth}, and (ii) employing a CNN as a refinement step, effectively removing blocky artifacts post-PCA reconstruction.

In this paper, we systematically investigate these patch-based PCA strategies on a 2D Poisson equation dataset, analyzing the trade-offs between computational complexity and reconstruction accuracy. Furthermore, extensive parametric studies are presented to evaluate the impact of patch size, overlap degree, and CNN refinement parameters, thereby offering practical insights and guidelines for efficiently employing patch-based PCA-Net neural operators.

\section{Methodology}

\subsection{2D Poisson's Dataset}

\begin{figure}
    \centering
    \includegraphics[width=1\textwidth]{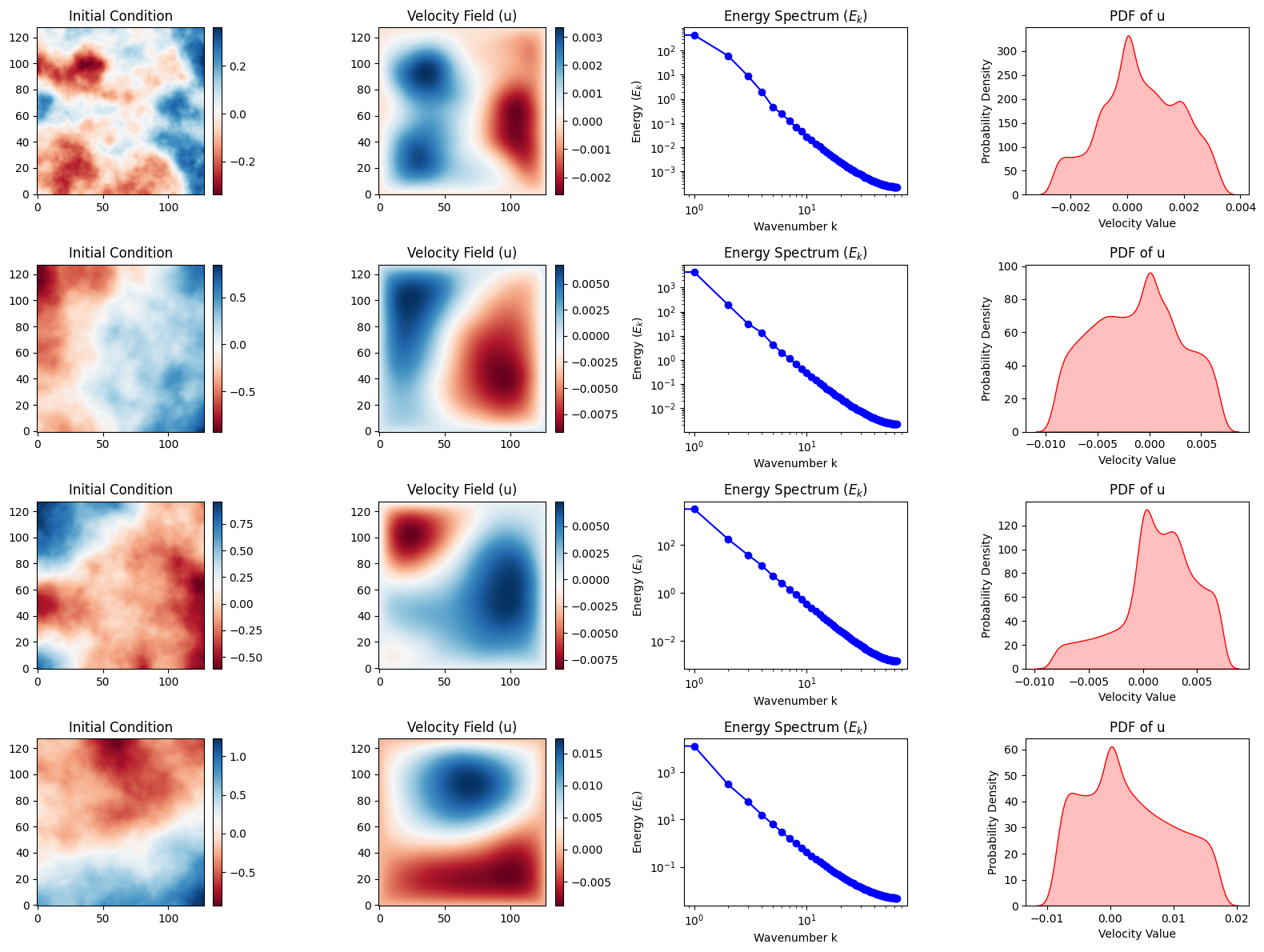}
    \caption{Visualization of four initial conditions and their corresponding solution fields, energy spectra, and probability density functions. The columns represent: (1) Initial Condition, (2) Velocity Field Solution $u$, (3) Energy Spectrum $E_k$, and (4) Probability Density Function (PDF). Each row corresponds to a different sample.}
    \label{fig:data_samples}
\end{figure}

We generate a synthetic dataset (see Fig. \ref{fig:data_samples} for some samples) representing solutions to the two-dimensional (2D) Poisson equation, widely used in scientific computing to model steady-state diffusion processes, electrostatics, and heat conduction problems \cite{huang2024diffusionpdegenerativepdesolvingpartial}. The Poisson equation is mathematically expressed as:

\begin{equation}
\nabla^2 u(x,y) = f(x,y), \quad (x,y) \in [0,1]^2,
\end{equation}
subject to homogeneous Dirichlet boundary conditions:
\begin{equation}
u|_{\partial \Omega} = 0,\end{equation}
where $u(x,y)$ is the solution field and $f(x,y)$ is a known source term or coefficient.

To construct the dataset, we generate $N = 10,000$ distinct realizations of the coefficient fields $f$ on a uniform $128 \times 128$ grid. Each realization of $f$ is synthesized using a Gaussian Random Field (GRF) parameterized by spectral characteristics governed by parameters $\alpha = 3$ and $\tau = 3$, as shown below:

\begin{equation}
f \sim \text{GRF}(\alpha=3, \tau=3).
\end{equation}

Once these coefficient fields are obtained, the corresponding solution fields $u(x,y)$ are computed by discretizing and solving the Poisson equation:
\begin{equation}
\nabla^2 u(x,y) = f(x,y), \quad \text{on grid } S \times S,
\end{equation}
using a finite difference discretization with a standard 5-point Laplacian stencil. Specifically, we discretize the spatial domain $\Omega=[0,1]^2$ onto a uniform grid of resolution $128 \times 128$. The discrete Poisson equation can then be represented as a linear system:

\begin{equation}
A\mathbf{u} = b,
\end{equation}
where $\mathbf{u}\in\mathbb{R}^{16384}$ is the flattened vector representation of the solution field $u$, and $b \in \mathbb{R}^{16384}$ is the corresponding vectorized source term $f(x,y)$. The matrix $A$ represents the discretized Laplacian operator, structured as:

\begin{equation}
A_{ij} =
\begin{cases}
4, & \text{if } i=j\\
-1, & \text{if nodes } i \text{ and } j \text{ are adjacent neighbors}\\
0, & \text{otherwise},
\end{cases}
\end{equation}
subject to homogeneous Dirichlet boundary conditions enforced explicitly by setting boundary rows in $A$ equal to identity and corresponding entries in $f$ equal to zero.

We generate a total of 10,000 independent pairs of coefficient fields $f$ and solutions $u$, thereby constructing a comprehensive dataset for training and evaluating our neural operators. This dataset forms the basis for assessing the efficacy of our proposed patch-based PCA-Net methods, allowing thorough exploration of trade-offs between computational complexity, accuracy, and reconstruction quality.

\subsection{Patch-based PCA Approaches}

\begin{figure}
    \centering
    \includegraphics[width=0.9\textwidth]{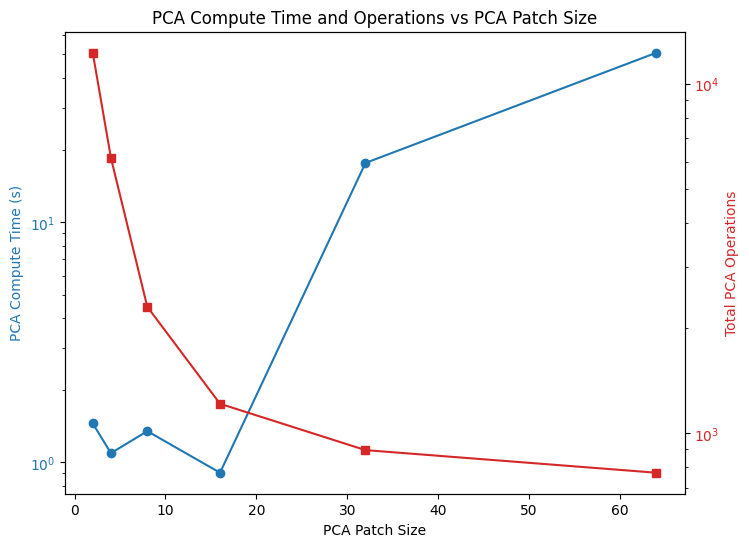}
    \caption{Comparison of PCA compute time and number of operations across different patch sizes. The x-axis represents the PCA patch size, while the y-axis shows the computational time (left) and the total number of PCA operations (right). The log-log scale highlights the trade-off between computational efficiency and accuracy, illustrating how smaller patches reduce PCA compute overhead but increase the total number of operations due to a larger number of patches.}
    \label{fig:pca_patch_size}
\end{figure}

The core idea behind patch-based PCA approaches is to partition high-dimensional fields into smaller patches to substantially reduce computational complexity associated with global PCA operations (see Fig. \ref{fig:pca_patch_size}). Specifically, given a field discretized onto a $D \times D$ grid, each field is partitioned into smaller patches, each of size $p \times p$, resulting in:

\begin{equation}
P = \frac{D^2}{p^2},
\end{equation}
patches per sample, where typically $p \ll D$. Let $a_{p}^{(i)} \in \mathbb{R}^{p^2}$ denote the vectorized representation of the $p$-th patch from the $i$-th sample. PCA is independently performed within these patches, identifying reduced-dimensional representations by selecting principal components that retain a predefined variance ratio (e.g., 99\%). The localized PCA decomposition for patch $p$ is mathematically expressed as:

\begin{equation}
X_p = \left[ a_p^{(1)}, a_p^{(2)}, \dots, a_p^{(m)} \right]^T = U_p \Sigma_p V_p^T,
\end{equation}
where $m$ is the number of training samples, and $U_p$, $\Sigma_p$, and $V_p$ denote singular value decomposition (SVD) components for the patch data.

\subsubsection{Approach A: Local-to-Global patch mapping}

In the Local-to-Global patch mapping approach, PCA is independently applied to each localized patch of the input coefficient fields, while the output solution fields are handled globally with a single PCA operation. This preserves global coherence in the solution fields while keeping the input representation compact.

Let $P$ be the (non-overlapping) patch count per sample and $k_p\ll p^2$ the number of retained modes for input patch $p$. For the $i$-th sample we stack all local latent codes into one vector
\begin{equation}
  \mathbf{x}^{(i)}
      \;=\; \bigl[\,
        \mathbf{x}_1^{(i)},\,
        \mathbf{x}_2^{(i)},\,
        \dots,\,
        \mathbf{x}_P^{(i)}
      \bigr]
      \;\in\;\mathbb{R}^{k_{\mathrm{in}}}, 
  \qquad
  k_{\mathrm{in}}=\sum_{p=1}^{P} k_p,
\end{equation}
with $\mathbf{x}_p^{(i)}\in\mathbb{R}^{k_p}$. 
\emph{(Input encoding.)} For each input patch $p$, let $\mathbf{F}_{p,\mathrm{raw}}\in\mathbb{R}^{m\times p^2}$ collect the patch pixels row-wise over $m$ samples, let $\boldsymbol{\mu}_{p,\mathrm{in}}\in\mathbb{R}^{p^2}$ be the column mean, and define the centered matrix $\mathbf{F}_p=\mathbf{F}_{p,\mathrm{raw}}-\mathbf{1}_m\boldsymbol{\mu}_{p,\mathrm{in}}^{\!\top}$. With the SVD $\mathbf{F}_p=\tilde{\mathbf{U}}_{p}\tilde{\boldsymbol{\Sigma}}_{p}\tilde{\mathbf{V}}_{p}^{\!\top}$, the patch basis is $\boldsymbol{\Psi}_{p,\mathrm{in}} \equiv \tilde{\mathbf{V}}_{p,1:k_p}\in\mathbb{R}^{p^2\times k_p}$ and
\begin{equation}
  \mathbf{x}_p^{(i)} \;=\; \boldsymbol{\Psi}_{p,\mathrm{in}}^{\!\top}\bigl(\mathbf{f}_p^{(i)}-\boldsymbol{\mu}_{p,\mathrm{in}}\bigr),
\end{equation}
where $\mathbf{f}_p^{(i)}\in\mathbb{R}^{p^2}$ is the raw input patch of sample $i$.

\paragraph{Output side (global PCA).}
Collect the raw solution fields (vectorized $n\times n$ images) as rows:
\begin{equation}
  \mathbf{U}_{\mathrm{raw}}
    \;=\;
    \begin{bmatrix}
      (\mathbf{u}^{(1)})^{\!\top}\\ \vdots \\ (\mathbf{u}^{(m)})^{\!\top}
    \end{bmatrix}
    \;\in\;\mathbb{R}^{m\times n^{2}}.
\end{equation}
Let $\boldsymbol{\mu}_{\mathrm{out}}\in\mathbb{R}^{n^2}$ be the column mean and define the centered matrix
$\mathbf{U}=\mathbf{U}_{\mathrm{raw}}-\mathbf{1}_m\boldsymbol{\mu}_{\mathrm{out}}^{\!\top}$.
Compute the SVD $\mathbf{U}=\tilde{\mathbf{U}}\tilde{\boldsymbol{\Sigma}}\tilde{\mathbf{V}}^{\!\top}$ and retain the top $k_{\mathrm{out}}\ll n^2$ right singular vectors
\begin{equation}
  \boldsymbol{\Phi}_{\mathrm{out}} \;\equiv\; \tilde{\mathbf{V}}_{1:k_{\mathrm{out}}}
  \;\in\;\mathbb{R}^{n^2\times k_{\mathrm{out}}}.
\end{equation}
Then the \emph{global} latent code for sample $i$ is obtained by an orthogonal projection onto this basis,
\begin{equation}
  \mathbf{y}^{(i)} \;=\; \boldsymbol{\Phi}_{\mathrm{out}}^{\!\top}\bigl(\mathbf{u}^{(i)}-\boldsymbol{\mu}_{\mathrm{out}}\bigr)\in\mathbb{R}^{k_{\mathrm{out}}}.
\end{equation}

\paragraph{Learned operator and decoding.}
A neural operator $F_{\boldsymbol\theta}$ maps the concatenated local input code to the global output code,
\begin{equation}
  \boxed{~~
    \mathbf{y}^{(i)} \;=\; F_{\boldsymbol\theta}\!\bigl(\mathbf{x}^{(i)}\bigr)
  ~~}
\end{equation}
and decoding to the physical field is a linear synthesis in the orthonormal basis:
\begin{equation}
  \hat{\mathbf{u}}^{(i)}
      \;=\; \boldsymbol{\mu}_{\mathrm{out}}
        + \boldsymbol{\Phi}_{\mathrm{out}}\,\mathbf{y}^{(i)}.
\end{equation}

\subsubsection{Approach B: Local-to-Local patch mapping}

In the Local-to-Local patch mapping approach, PCA is applied independently on each patch for both input coefficient fields and output solution fields. While significantly reducing computational complexity, this approach introduces challenges with reconstruction smoothness and coherence, particularly at patch boundaries.

\begin{figure}
    \centering
    \includegraphics[width=1\linewidth]{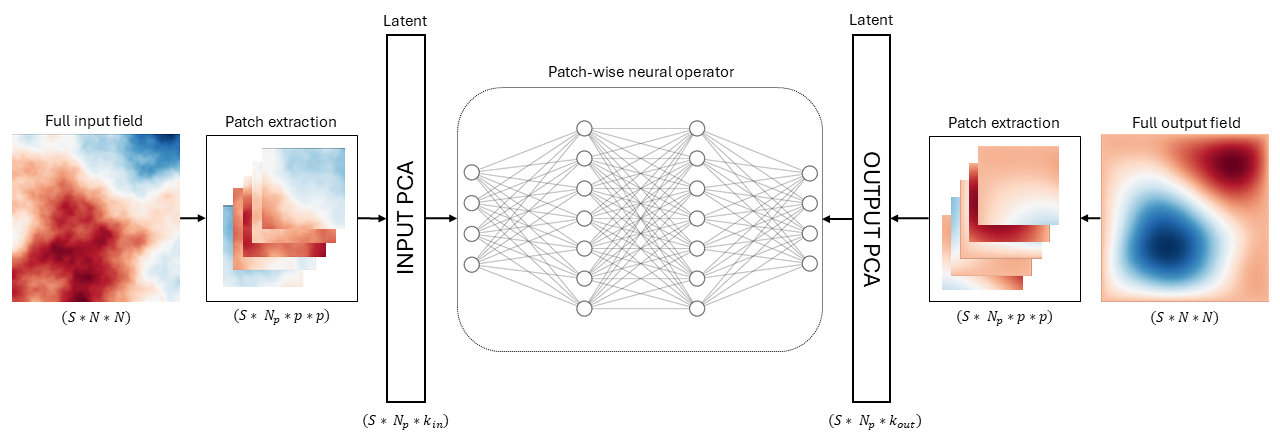}
    \caption{Model setup for the Local-to-local patch based PCA approach.}
    \label{fig:enter-label-1}
\end{figure}

For each \emph{output} patch $p$, collect pixels over $m$ samples row-wise:
\begin{equation}
  \mathbf{Y}_{p,\mathrm{raw}}
    \;=\;
    \begin{bmatrix}
      (\mathbf{u}_{p}^{(1)})^{\!\top}\\ \vdots \\ (\mathbf{u}_{p}^{(m)})^{\!\top}
    \end{bmatrix}
    \in\mathbb{R}^{m\times s^{2}}.
\end{equation}
Let $\boldsymbol{\mu}_{p,\mathrm{out}}\in\mathbb{R}^{p^2}$ be the column mean and define the centered matrix
$\mathbf{Y}_p=\mathbf{Y}_{p,\mathrm{raw}}-\mathbf{1}_m\boldsymbol{\mu}_{p,\mathrm{out}}^{\!\top}$.
With the SVD $\mathbf{Y}_p=\tilde{\mathbf{U}}_{p}\tilde{\boldsymbol{\Sigma}}_{p}\tilde{\mathbf{V}}_{p}^{\!\top}$,
retain $l_p\ll p^2$ right singular vectors
$\boldsymbol{\Phi}_{p,\mathrm{out}}\equiv\tilde{\mathbf{V}}_{p,1:l_p}\in\mathbb{R}^{p^2\times l_p}$ and encode
\begin{equation}
  \mathbf{y}_{p}^{(i)} \;=\; \boldsymbol{\Phi}_{p,\mathrm{out}}^{\!\top}\bigl(\mathbf{u}_p^{(i)}-\boldsymbol{\mu}_{p,\mathrm{out}}\bigr)
  \;\in\;\mathbb{R}^{l_p}.
\end{equation}

\paragraph{Concatenated feature vectors.}
\begin{align}
  \mathbf{x}^{(i)} &=
      \bigl[\mathbf{x}_1^{(i)},\dots,\mathbf{x}_P^{(i)}\bigr]
      \in\mathbb{R}^{k_{\mathrm{in}}}, 
  \qquad
  k_{\mathrm{in}}=\sum_{p=1}^{P}k_p,\\[2pt]
  \mathbf{y}^{(i)} &=
      \bigl[\mathbf{y}_1^{(i)},\dots,\mathbf{y}_P^{(i)}\bigr]
      \in\mathbb{R}^{l_{\mathrm{out}}},
  \qquad
  l_{\mathrm{out}}=\sum_{p=1}^{P}l_p.
\end{align}

\paragraph{Learned operator and decoding.}
\begin{equation}
  \boxed{~~
    \mathbf{y}^{(i)} = F_{\boldsymbol\theta}\!\bigl(\mathbf{x}^{(i)}\bigr)
  ~~}
\end{equation}
During inference, each predicted patch code is decoded via
\begin{equation}
  \hat{\mathbf{u}}_{p}^{(i)} \;=\; \boldsymbol{\mu}_{p,\mathrm{out}} + \boldsymbol{\Phi}_{p,\mathrm{out}}\,\mathbf{y}_{p}^{(i)}.
\end{equation}
During inference, $\mathbf{y}_p^{(i)}$ are decoded with the local
basis~$(\tilde{\mathbf{U}}_p,\tilde{\boldsymbol{\Sigma}}_p,\tilde{\mathbf{V}}_p)$
and the reconstructed patches are mosaicked—optionally with overlap–
additive blurring or a post-hoc CNN refinement—to yield the final
$\,\hat{u}(x,y)\,$ field.

\subsection{Patch Overlap and Smoothing}

To address reconstruction artifacts at patch boundaries inherent to the local-to-local PCA approach, we introduce an overlapping patch strategy combined with smoothing filters. Rather than using strictly non-overlapping partitions, each solution field is segmented into patches that partially overlap. This overlapping is controlled by selecting a stride length $s$ smaller than the patch size $p$ ($s < p$), effectively increasing the number of patches and enhancing continuity at the expense of additional computational complexity.

To achieve smooth transitions between overlapping patches during reconstruction, we employ a smoothing window, specifically the Hanning window. The one-dimensional Hanning window function is defined as:

\begin{equation}
\text{Hanning}(n) = 0.5 \left[ 1 - \cos\left(\frac{2\pi n}{p+1}\right) \right], \quad n \in {1,2,\dots,p}.
\end{equation}

The two-dimensional smoothing window $W \in \mathbb{R}^{p \times p}$ is then obtained by taking the outer product of two one-dimensional Hanning windows:

\begin{equation}
W(i,j) = \text{Hanning}(i) \times \text{Hanning}(j), \quad i,j \in {1,2,\dots,p}.
\end{equation}

During the assembly of the full solution field from individually reconstructed patches, each patch is multiplied element-wise by this smoothing window. Overlapping regions between adjacent patches are then averaged, producing solution fields with significantly reduced boundary discontinuities and improved visual and numerical continuity.

\subsection{CNN based RefinementNet}

As a complementary approach to reduce boundary artifacts, we employ a CNN refinement stage, termed RefinementNet. This CNN operates directly on reconstructed solution fields obtained from the local-to-local PCA model. The RefinementNet architecture is designed as a shallow CNN comprising multiple convolutional layers followed by nonlinear activation functions. Formally, the refinement step can be expressed as:

\begin{equation}
\hat{u} = \sigma\left(W_{N} * \sigma\left(W_{N-1} * \dots \sigma\left(W_{1} * \tilde{u} + b_{1}\right) \dots + b_{N-1}\right) + b_{N}\right),
\end{equation}

where $\tilde{u}$ denotes the blocky reconstruction obtained from local PCA patches, $\hat{u}$ is the refined solution, $W_i$ and $b_i$ represent the convolutional kernels and biases for the $i^{th}$ layer, respectively, and $\sigma$ is the nonlinear activation function, the Rectified Linear Unit (ReLU).
By varying the convolution kernel sizes (e.g., $3\times3$, $5\times5$, $7\times7$), we investigate the effect of CNN receptive fields on the refinement performance. The CNN-based refinement effectively reduces residual artifacts and yields smoother, higher-quality reconstructions without significantly increasing computational overhead.

\section{Experimentation \& Results}

\begin{figure}
    \centering
    \includegraphics[width=1\linewidth]{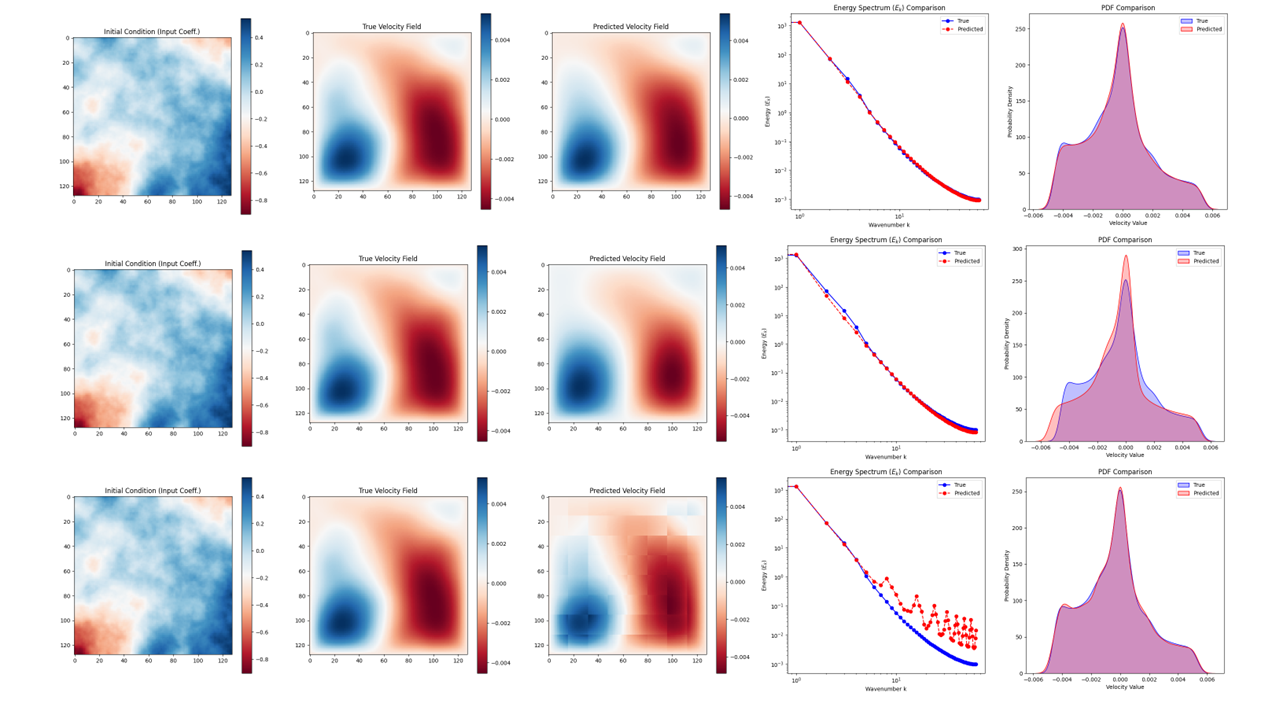}
    \caption{Patch-PCA approach comparison with the ground truth}
    \label{fig:patch_comp_with_gt}
\end{figure}

Fig. \ref{fig:patch_comp_with_gt} presents a comparison of the reconstruction quality across the three PCA-based neural operator approaches: Global PCA, Local-to-Global PCA, and Local-to-Local PCA. The figure is organized into three rows, each corresponding to one of these approaches. Each row consists of five columns displaying, respectively: (1) the initial conditions, (2) the ground truth solution field $u$, (3) the predicted or reconstructed solution field $u$, (4) a comparison of energy spectra $E_k$, and (5) a comparison of probability density functions (PDF).

The visual analysis of the reconstructed fields clearly indicates that the Global PCA and Local-to-Global PCA approaches produce smooth and visually coherent fields without noticeable blockiness. In contrast, the Local-to-Local PCA method exhibits distinct blocky artifacts, primarily at patch boundaries. These artifacts arise due to the independent patch-wise processing of the output fields.

Examining the energy spectra ($E_k$), both Global PCA and Local-to-Global PCA reconstructions closely align with the ground truth spectrum across all wavenumbers, demonstrating their ability to preserve the spectral characteristics of the original fields accurately. Conversely, the Local-to-Local PCA approach shows alignment with the ground truth at lower wavenumbers but diverges at higher wavenumbers, evident from the pronounced zig-zag patterns. This behavior reflects difficulties in preserving global spectral coherence when using purely localized PCA reconstructions.

Further insights are provided by the probability density function (PDF) comparisons. Both Global PCA and Local-to-Local PCA methods show excellent alignment with the ground truth PDFs, indicating effective preservation of the statistical distribution of the solution fields. The Local-to-Global PCA, however, exhibits a slightly shifted PDF compared to the ground truth, suggesting minor deviations in capturing the statistical distributions.

Quantitative performance metrics summarized in Table \ref{tab:model_comparison} reinforce these observations. The Global PCA approach achieves the lowest mean squared error (MSE = $1.500 \times 10^{-8}$) and mean absolute error (MAE = $9.195 \times 10^{-5}$), along with the highest structural similarity index (SSIM = $0.9963$). The Local-to-Global PCA has higher errors (MSE = $1.010 \times 10^{-7}$, MAE = $2.360 \times 10^{-4}$) and lower SSIM ($0.9801$), reflecting a slight compromise in reconstruction accuracy. The Local-to-Local PCA method, despite exhibiting visual blockiness, achieves competitive quantitative metrics (MSE = $6.000 \times 10^{-8}$, MAE = $1.468 \times 10^{-4}$, SSIM = $0.9926$), highlighting the trade-off between visual coherence and numerical accuracy.

\begin{table}
    \centering
    \small
    \begin{tabular}{lcccc}
        \toprule
        \textbf{Model} & \textbf{No. of Parameters} & \textbf{MSE} & \textbf{MAE} & \textbf{SSIM} \\
        \midrule
        Global PCA & 128022 & $1.500 \times 10^{-8}$ & $9.195 \times 10^{-5}$ & 0.9963 \\
        Local-to-Global PCA & 64138 & $1.010 \times 10^{-7}$ & $2.360 \times 10^{-4}$ & 0.9801 \\
        Local-to-Local PCA & 79876 & $6.000 \times 10^{-8}$ & $1.468 \times 10^{-4}$ & 0.9926 \\
        \bottomrule
    \end{tabular}
    \caption{Performance comparison of different PCA-based neural operator models. MSE: Mean Squared Error, MAE: Mean Absolute Error, SSIM: Structural Similarity Index.}
    \label{tab:model_comparison}
\end{table}

\begin{figure}
    \centering
    \includegraphics[width=1\linewidth]{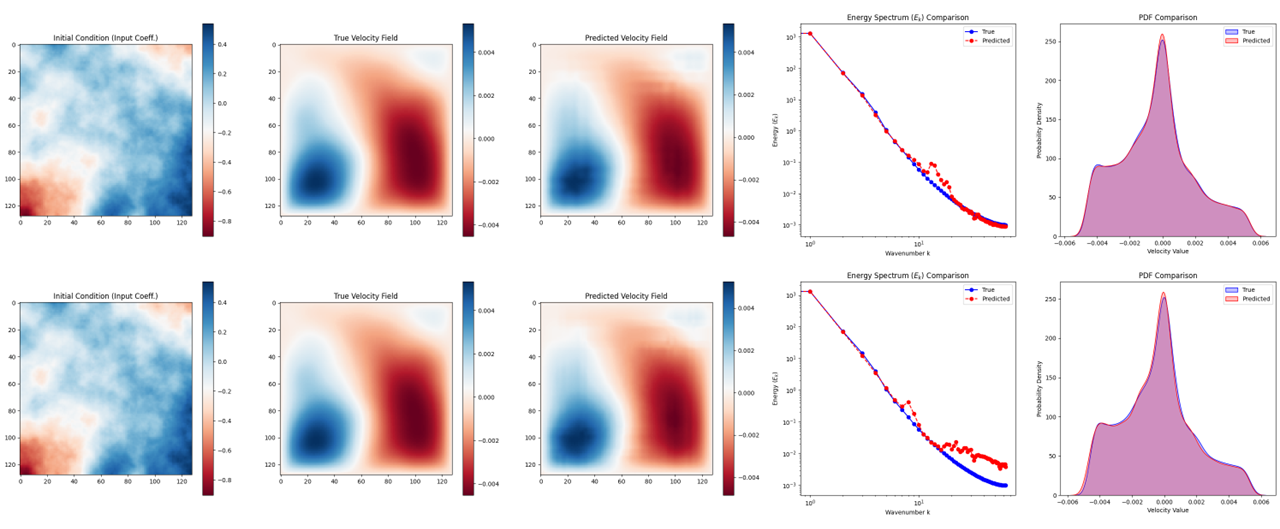}
    \caption{Top: Patch PCA, bottom: RefinementNet PCA}
    \label{fig:aug_model_comp}
\end{figure}

Fig. \ref{fig:aug_model_comp} presents a comparison between the refinements applied to the Local-to-Local PCA approach, specifically the Patch Overlap with smoothing and the RefinementNet CNN approaches. The figure consists of two rows, each corresponding to one of these refinements. Each row contains five columns displaying: (1) the initial conditions, (2) the ground truth solution field $u$, (3) the reconstructed solution field $u$, (4) the energy spectra $E_k$ comparison, and (5) the probability density function (PDF) comparison.

Visual analysis of the reconstructed solution fields demonstrates a significant reduction in blocky artifacts for both refinement strategies compared to the standard Local-to-Local PCA approach. Specifically, the Patch Overlap with smoothing filter effectively mitigates boundary discontinuities, producing smooth and coherent solution fields. Similarly, the RefinementNet CNN approach notably reduces artifacts, resulting in visually improved reconstructions.

Further examination of the energy spectra ($E_k$) plots reveals that the Patch Overlap approach aligns closely with the ground truth across all wavenumbers, effectively preserving global spectral characteristics. The RefinementNet CNN approach shows good alignment at lower wavenumbers but exhibits slight deviations at higher wavenumbers, suggesting minor limitations in capturing high-frequency spectral details accurately.

The probability density function comparisons demonstrate excellent agreement for both refinements with the ground truth distributions, highlighting their effectiveness in preserving the statistical properties of the original solution fields.

Quantitative metrics summarized in Table \ref{tab:l2l_app_model_comparison} further reinforce these findings. The Patch Overlap refinement achieves superior performance (MSE = $1.000 \times 10^{-8}$, MAE = $6.855 \times 10^{-5}$, SSIM = $0.9984$) compared to the RefinementNet CNN (MSE = $3.100 \times 10^{-8}$, MAE = $1.269 \times 10^{-4}$, SSIM = $0.9910$), surpassing even the Global PCA method. However, the Patch Overlap refinement incurs a higher parameter count (202,323 parameters) compared to the RefinementNet CNN approach (98,981 parameters), indicating a trade-off between computational complexity and reconstruction accuracy.

\begin{table}
    \centering
    \begin{tabular}{lcccc}
        \toprule
        \textbf{Model} & \textbf{No. of Parameters} & \textbf{MSE} & \textbf{MAE} & \textbf{SSIM} \\
        \midrule
        Global PCA & 128022 & $1.500 \times 10^{-8}$ & $9.195 \times 10^{-5}$ & 0.9963 \\
        Patch Overlap & 202323 & $1.000 \times 10^{-8}$ & $6.855 \times 10^{-5}$ & 0.9984 \\
        RefinementNet & 98981 & $3.100 \times 10^{-8}$ & $1.269 \times 10^{-4}$ & 0.9910 \\
        \bottomrule
    \end{tabular}
    \caption{Performance comparison of different PCA-based neural operator models. MSE: Mean Squared Error, MAE: Mean Absolute Error, SSIM: Structural Similarity Index.}
    \label{tab:l2l_app_model_comparison}
\end{table}

\begin{figure}
    \centering
    \includegraphics[width=1\linewidth]{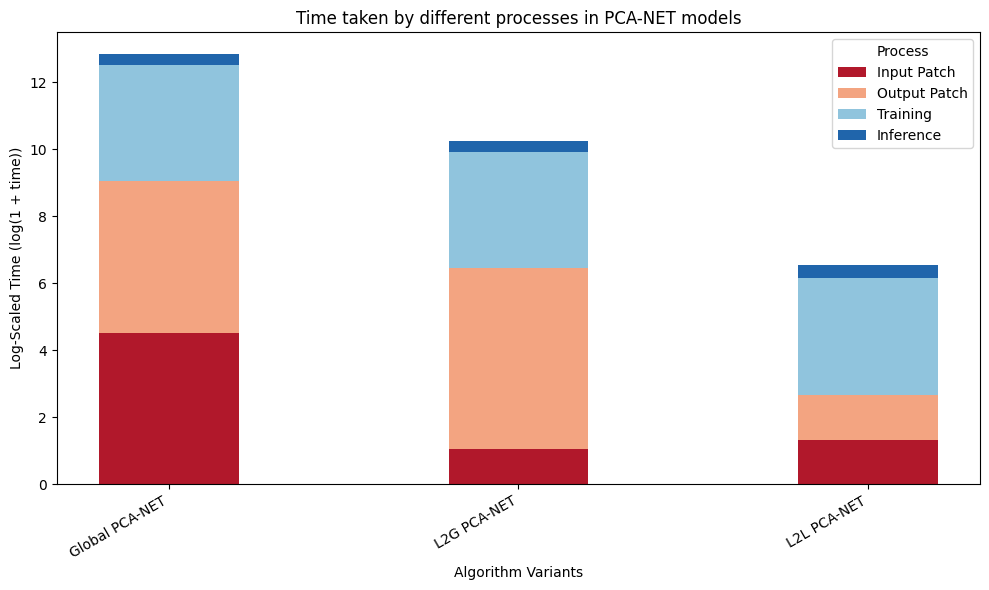}
    \caption{Log-Scaled Time comparison for various processes in the PCA-Net models for 1) Global PCA-Net, 2) Local-to-Global PCA-Net, and 3) Local-to-Local PCA-Net.}
    \label{fig:3var_time_comp}
\end{figure}

Fig. \ref{fig:3var_time_comp} presents a detailed histogram comparison illustrating the computational time (log-scaled) required for distinct processes—input patch PCA processing (red), output patch PCA processing (beige), model training (light blue), and inference (dark blue)—across the three PCA-Net variants: Global PCA-Net, Local-to-Global PCA-Net, and Local-to-Local PCA-Net.

Analysis of this figure clearly indicates that the Global PCA-Net consumes the most significant computational resources, primarily attributed to the extensive time required for input and output PCA computations due to the large dimensionality of the global fields. The Local-to-Global PCA-Net variant shows modest improvements in reducing these PCA processing times, reflecting partial alleviation of computational burdens by introducing local PCA computations for inputs.

Notably, the Local-to-Local PCA-Net variant achieves the most substantial reduction in overall computational time, drastically minimizing input and output PCA processing durations. Although this variant experiences a slight increase in model training time, resulting from processing multiple patches per sample, this increase is minimal compared to the significant savings realized in PCA computation. Consequently, the entire pipeline—from PCA processing to inference—is significantly more efficient when employing the Local-to-Local PCA-Net approach compared to the other two variants. In the upcoming experimentation, we will see results from the modified versions of the Local-to-Local PCA-Net.  

\begin{figure}
    \centering
    \includegraphics[width=1\linewidth]{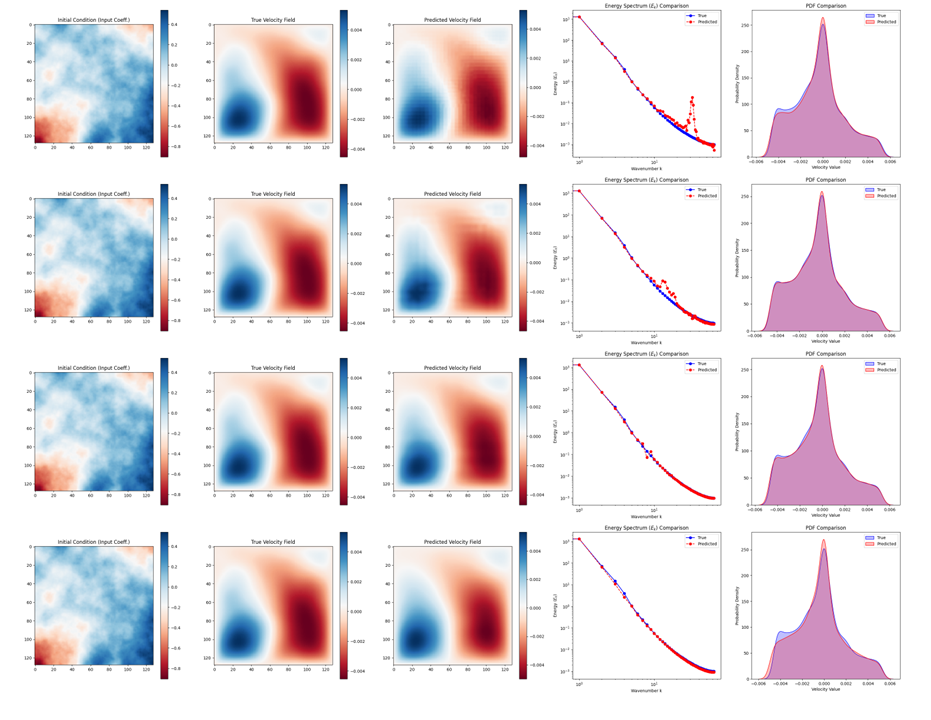}
    \caption{Row 1: Patch size = 8, Stride = 4, Row 2: Patch size = 16, Stride = 8, Row 3: Patch size = 32, Stride = 16, Row 4: Patch size = 64, Stride = 32}
    \label{fig:parametric_vis_overlap}
\end{figure}

Fig. \ref{fig:parametric_vis_overlap} illustrates the effectiveness of overlapping patches for four different patch sizes: $8\times 8$, $16\times 16$, $32\times 32$, and $64\times 64$. Each row in the figure corresponds to one of these patch sizes, with overlaps of 4, 8, 16, and 32 pixels respectively. Within each row, five columns are shown: (1) the initial condition, (2) the ground-truth solution, (3) the reconstructed or predicted solution, (4) the energy spectrum $E_k$, and (5) PDF. Visually, all four patch sizes yield coherent solution fields with substantially reduced blockiness, confirming the utility of overlapping patches in mitigating reconstruction artifacts. At larger patch sizes (e.g., $64\times 64$), the energy spectrum $E_k$ aligns even more closely with the ground truth, reflecting an enhanced ability to capture higher-wavenumber content. Minor deviations are observed in the PDF plots for some configurations, particularly in the $64\times 64$ case, which suggests that although overlap alleviates discontinuities, there remains a trade-off between patch dimension, overlap size, and exact statistical alignment with the ground truth.

A more extensive parametric investigation of overlapping patches is presented in Table \ref{tab:patch_size_comparison}, where we vary the stride size (i.e., the degree of overlap) for each patch dimension. In this table, we report the number of patches extracted, the total extraction time, and the resulting reconstruction metrics (MSE, MAE, and SSIM). Smaller patch sizes yield lower PCA costs per patch but dramatically increase the total number of patches that must be processed, thereby raising the cumulative extraction time. Conversely, larger patches reduce the overall patch count but increase the dimensionality of each patch-based PCA and potentially its associated compute time. Hence, the choice of patch size and stride warrants a balance between accuracy and efficiency. From an error metrics perspective, many of the examined configurations achieve comparably low MSE and MAE values, along with high SSIM scores. However, as the table shows, some settings are more computationally intensive than others due to the interplay between patch size, overlap, and the total number of patches. Consequently, identifying a suitable “sweet spot” for both accuracy and computational cost is essential—one that avoids excessive overhead while maintaining high reconstruction fidelity.

\begin{table}
    \centering
    \renewcommand{\arraystretch}{1.2} 
    \resizebox{\textwidth}{!}{
    \begin{tabular}{c|c|c|c|c|c|c}
        \toprule
        \textbf{Patch Size} & \textbf{Stride Size} & \textbf{No. of Patches Extracted} & \textbf{Time Taken (s)} & \textbf{MSE} & \textbf{MAE} & \textbf{SSIM} \\
        \midrule
        \multirow{3}{*}{8}  & 3  & 1681 & 31.0 & $5.55 \times 10^{-9}$ & $5.54 \times 10^{-5}$ & 0.998 \\
                             & 4  & 961  & 16.7 & $9.58 \times 10^{-9}$ & $7.06 \times 10^{-5}$ & 0.998 \\
                             & 5  & 625  & 12.0 & $2.63 \times 10^{-8}$ & $1.10 \times 10^{-4}$ & 0.995 \\
        \midrule
        \multirow{4}{*}{16} & 6  & 361  & 11.7 & $6.51 \times 10^{-8}$ & $1.14 \times 10^{-4}$ & 0.989 \\
                             & 8  & 225  & 8.5  & $1.24 \times 10^{-8}$ & $7.96 \times 10^{-5}$ & 0.996 \\
                             & 10 & 144  & 5.9  & $3.07 \times 10^{-8}$ & $1.11 \times 10^{-4}$ & 0.995 \\
                             & 12 & 100  & 4.0  & $9.23 \times 10^{-8}$ & $1.67 \times 10^{-4}$ & 0.985 \\
        \midrule
        \multirow{3}{*}{32} & 12 & 81   & 6.8  & $9.81 \times 10^{-9}$ & $7.44 \times 10^{-5}$ & 0.997 \\
                             & 16 & 49   & 4.6  & $1.13 \times 10^{-8}$ & $7.87 \times 10^{-5}$ & 0.997 \\
                             & 20 & 25   & 3.2  & $1.68 \times 10^{-8}$ & $5.00 \times 10^{-4}$ & 0.817 \\
        \midrule
        \multirow{3}{*}{64} & 24 & 9    & 2.5  & $1.72 \times 10^{-6}$ & $6.00 \times 10^{-4}$ & 0.809 \\
                             & 32 & 9    & 2.9  & $7.27 \times 10^{-8}$ & $1.97 \times 10^{-4}$ & 0.985 \\
                             & 40 & 4    & 1.8  & $3.92 \times 10^{-6}$ & $9.80 \times 10^{-4}$ & 0.700 \\
        \bottomrule
    \end{tabular}
    }
    \caption{Performance comparison of different patch sizes and stride values for PCA-based neural operator models. MSE: Mean Squared Error, MAE: Mean Absolute Error, SSIM: Structural Similarity Index.}
    \label{tab:patch_size_comparison}
\end{table}

\begin{figure}
    \centering
    \includegraphics[width=1\linewidth]{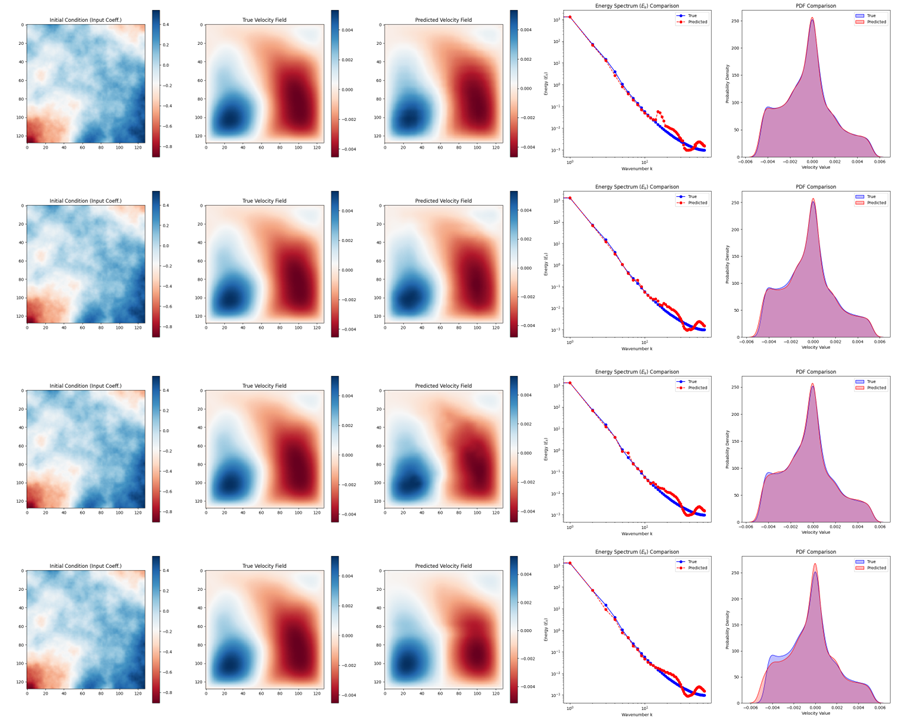}
    \caption{Parametric evaluation of RefinementNet kernel size, with each row corresponding to patch sizes of $8\times8$, $16\times16$, $32\times32$, and $64\times64$. }
    \label{fig:kernel_size_comp}
\end{figure}

Figure~\ref{fig:kernel_size_comp} presents a detailed examination of the RefinementNet’s kernel size configurations under varying patch sizes. The figure contains four rows corresponding to patch sizes of $8\times8$, $16\times16$, $32\times32$, and $64\times64$, each employing a $5\times5$ convolution kernel. Within each row, five columns are shown: (1) the initial condition, (2) the ground truth, (3) the reconstructed prediction after RefinementNet, (4) the energy spectrum $E_k$, and (5) the probability density function (PDF). Visually, the $16\times16$ patch size with a $5\times5$ kernel produces particularly coherent solutions with minimal patch-induced blockiness. Although the $E_k$ plots do not align quite as well with the ground truth as those from the overlap-based approach, they still represent a clear improvement over the unmodified Local-to-Local PCA method. The PDF plots align closely with the ground truth in most cases, with only minor deviations observed for the $64\times64$ patch size configuration.

Table~\ref{tab:patch_kernel_comparison} provides a more expansive parametric study, covering kernel sizes of $3\times3$, $5\times5$, and $7\times7$ across these four patch sizes. It also lists the total number of trainable parameters for each configuration alongside the corresponding error metrics (MSE, MAE, SSIM). The results show that while smaller patch sizes can yield marginally better error metrics, they often lead to a larger parameter count due to the increased number of patches (and, consequently, a more substantial RefinementNet). Conversely, larger patch sizes reduce the total patch count but may not capture finer-scale details as effectively. Across all patch sizes, a $5\times5$ kernel emerges as the most robust receptive field choice, striking a favorable balance between reconstruction accuracy and model complexity.

\begin{table}
    \centering
    \small 
    \renewcommand{\arraystretch}{1.2}
    \resizebox{\textwidth}{!}{ 
    \begin{tabular}{c|c|c|c|c|c}
        \toprule
        \textbf{Patch Size} & \textbf{Conv. Kernel} & \textbf{Parameters (PCA-Net + RefinementNet)} & \textbf{MSE} & \textbf{MAE} & \textbf{SSIM} \\
        \midrule
        \multirow{3}{*}{8}  & 3  & 164801  & $1.40 \times 10^{-8}$ & $9.256 \times 10^{-5}$ & 0.9943 \\
                             & 5  & 198593  & $1.00 \times 10^{-8}$ & $7.721 \times 10^{-5}$ & 0.9978 \\
                             & 7  & 249281  & $9.00 \times 10^{-9}$ & $7.403 \times 10^{-5}$ & 0.9979 \\
        \midrule
        \multirow{3}{*}{16} & 3  & 98981   & $3.10 \times 10^{-8}$ & $1.269 \times 10^{-4}$ & 0.9910 \\
                             & 5  & 132773  & $1.90 \times 10^{-8}$ & $1.006 \times 10^{-4}$ & 0.9962 \\
                             & 7  & 183461  & $2.00 \times 10^{-8}$ & $1.014 \times 10^{-4}$ & 0.9961 \\
        \midrule
        \multirow{3}{*}{32} & 3  & 76881   & $2.80 \times 10^{-8}$ & $1.243 \times 10^{-4}$ & 0.9932 \\
                             & 5  & 110673  & $2.10 \times 10^{-8}$ & $1.091 \times 10^{-4}$ & 0.9959 \\
                             & 7  & 161361  & $2.10 \times 10^{-8}$ & $1.099 \times 10^{-4}$ & 0.9957 \\
        \midrule
        \multirow{3}{*}{64} & 3  & 67889   & $1.11 \times 10^{-6}$ & $2.469 \times 10^{-4}$ & 0.9762 \\
                             & 5  & 101681  & $1.01 \times 10^{-6}$ & $2.347 \times 10^{-4}$ & 0.9806 \\
                             & 7  & 152369  & $1.01 \times 10^{-6}$ & $2.351 \times 10^{-4}$ & 0.9801 \\
        \bottomrule
    \end{tabular}
    }
    \caption{Performance comparison of different patch sizes and convolutional kernel sizes for PCA-Net and RefinementNet models. MSE: Mean Squared Error, MAE: Mean Absolute Error, SSIM: Structural Similarity Index.}
    \label{tab:patch_kernel_comparison}
\end{table}

\begin{figure}
    \centering
    \begin{subfigure}{0.75\textwidth}
        \centering
        \includegraphics[width=\textwidth]{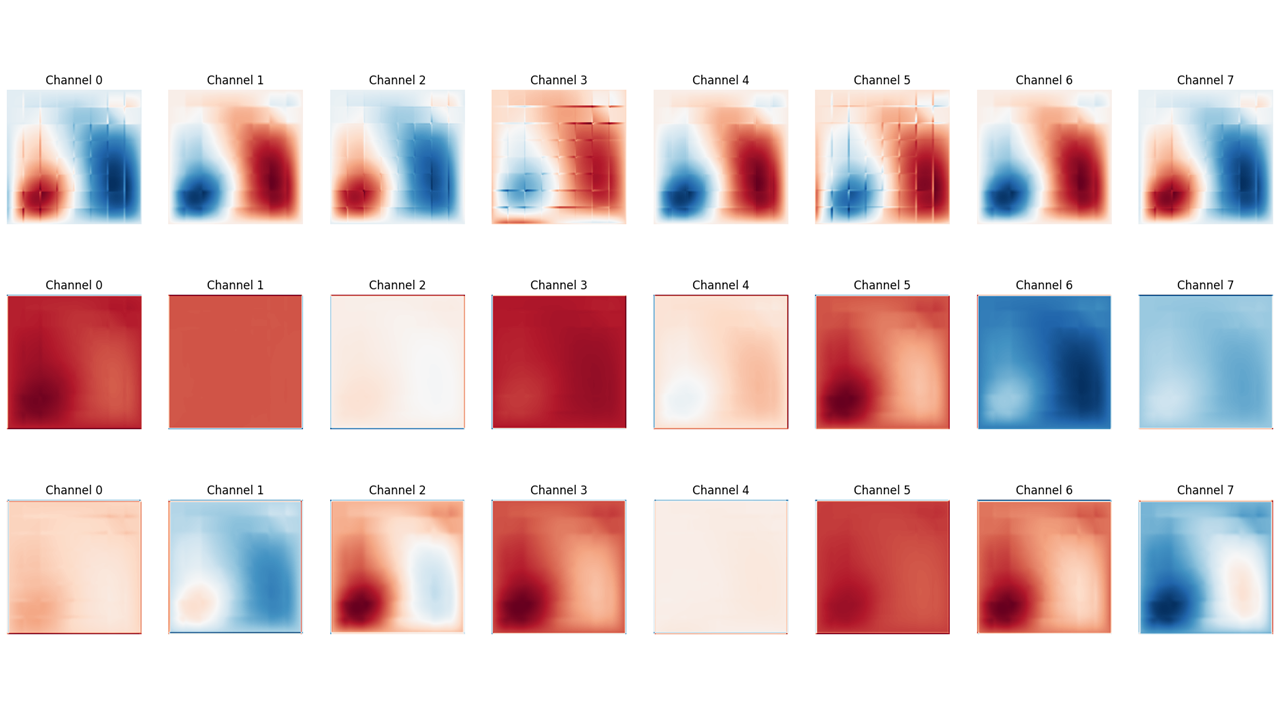}
        \caption{Kernel size = 3}
    \end{subfigure}
    \begin{subfigure}{0.75\textwidth}
        \centering
        \includegraphics[width=\textwidth]{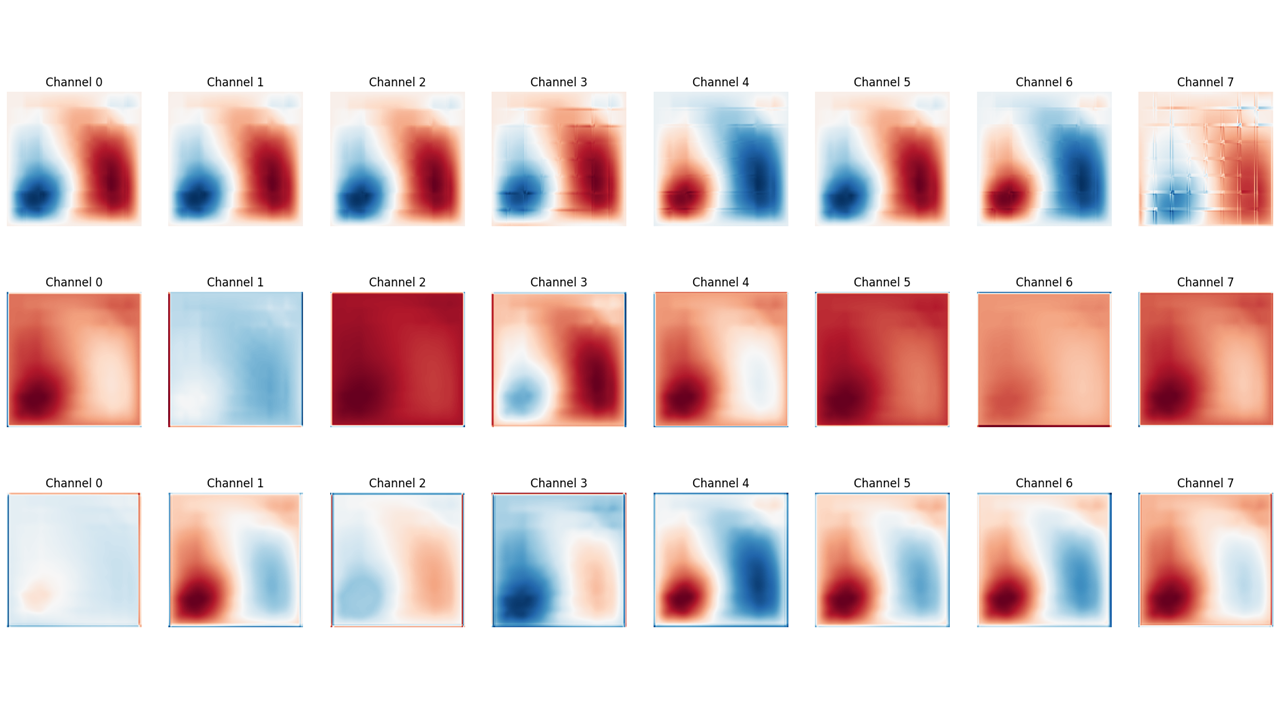}
        \caption{Kernel size = 5}
    \end{subfigure}
    
    
    \begin{subfigure}{0.75\textwidth}
        \centering
        \includegraphics[width=\textwidth]{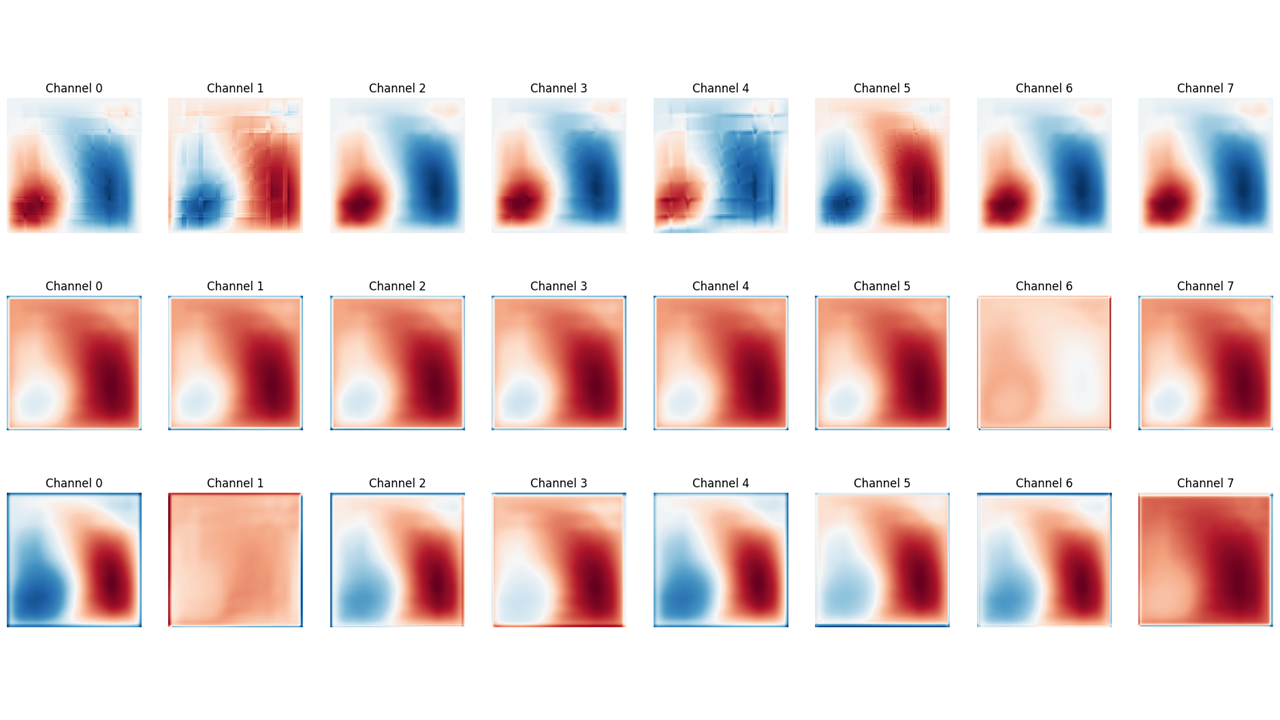}
        \caption{Kernel size = 7}
    \end{subfigure}
    \caption{RefinementNet receptive field ablation study showcasing sensitivities across three different kernel sizes.}
    \label{fig:refinementNet_recept_field}
\end{figure}

Fig. \ref{fig:refinementNet_recept_field} depicts feature-map ablation studies for the RefinementNet under three convolution kernel sizes: (a) $3\times3$, (b) $5\times5$, and (c) $7\times7$. Each subfigure presents three representative feature maps, corresponding to the initial, intermediate, and final layers of the network. As the activations propagate through these layers, the blocky artifacts and discontinuities become progressively attenuated. Notably, models employing larger kernels (i.e., $5\times5$ or $7\times7$) demonstrate more pronounced smoothing effects and reduced boundary inconsistencies, aligning with the improved quantitative metrics observed in earlier experiments.

\begin{figure}
    \centering
    \includegraphics[width=1\linewidth]{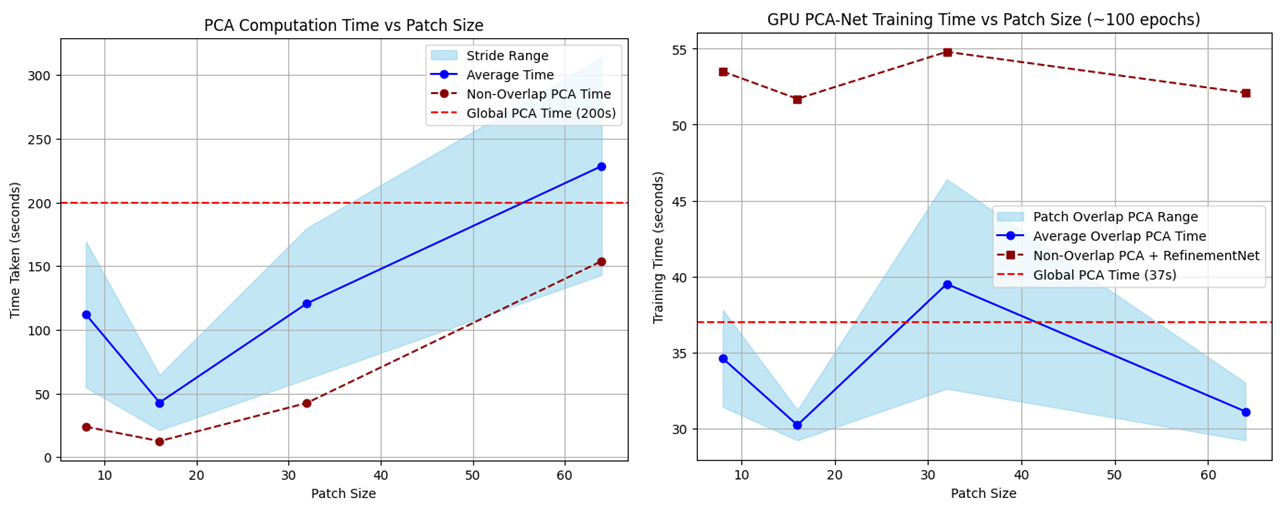}
    \caption{(Left) Time taken for the PCA compute. (Right) Time taken for the model training on a nvidia RTX4090 GPU.}
    \label{fig:appr_comp}
\end{figure}

Fig. \ref{fig:appr_comp} summarizes the overall computational times for various patch-based PCA approaches, with the left plot comparing PCA compute times and the right plot depicting GPU training times (over roughly 100\,epochs). In the PCA time plot, a dashed red line indicates the Global PCA compute time at around 200\,seconds, while the dark blue line and its light blue envelope show average and stride-range times for the Overlap PCA across each patch size. Notably, patch sizes of $8\times8$, $16\times16$, and $32\times32$ yield shorter compute times than the global baseline, whereas $64\times64$ slightly exceeds it due to higher per-patch dimensionality. In contrast, the dark red dashed line corresponds to Non-Overlap PCA, which remains below the overlap-based times for most patch sizes, with $16\times16$ exhibiting the lowest overall PCA compute cost. Turning to the GPU training plot, the dashed red line at approximately 37\,seconds marks the Global PCA training baseline. The dark blue line with light blue shading again corresponds to the overlap approach, showing lower training times than the global baseline for most patch sizes except $32\times32$. Meanwhile, the dark red dashed line represents the Non-Overlap PCA combined with RefinementNet, and it typically surpasses the global PCA training time. This arises because while non-overlap PCA offers initial efficiency gains, the additional complexity and parameters introduced by the RefinementNet can offset those savings during training.

\begin{figure}
    \centering
    \includegraphics[width=1\linewidth]{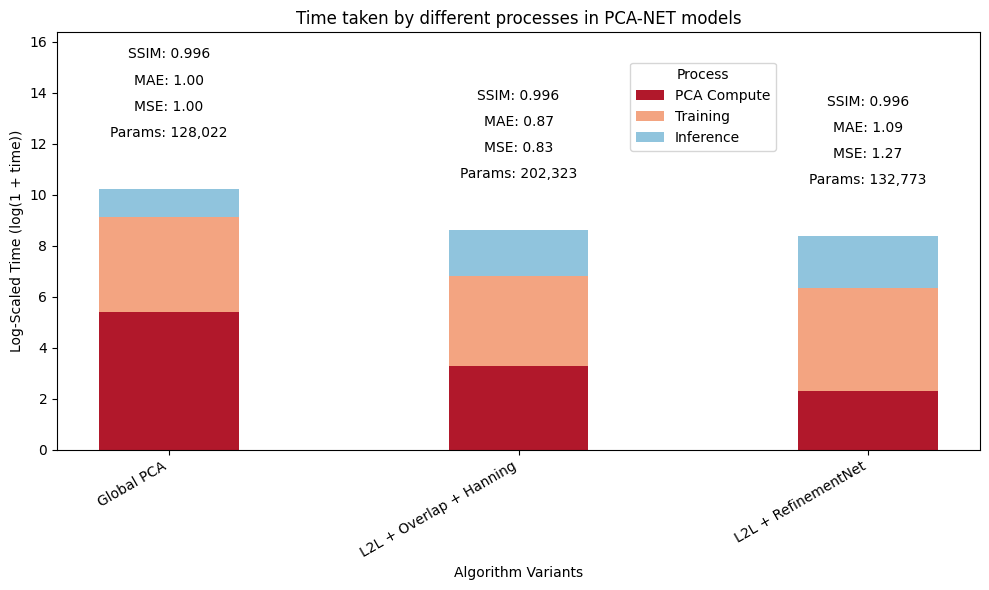}
    \caption{Comparison between the Local-2-local patch approach modifications.}
    \label{fig:final_time_comp}
\end{figure}

Fig. \ref{fig:final_time_comp} provides a log-scaled, stacked-bar visualization of the total computational time required by three PCA-Net variants: Global PCA, Local-to-Local (L2L) with overlapping patches, and L2L with RefinementNet. Each bar is subdivided by process type—PCA computation (red), training (beige), and inference (blue)—offering a clear breakdown of where computational resources are most heavily spent.

 Global PCA incurs the highest overall time, driven primarily by its expensive PCA step (red segment). The L2L + Overlap approach lowers the PCA overhead by localizing the dimensionality reduction, yet retains a moderate training cost (beige) because only a single network is being trained. In contrast, L2L + RefinementNet achieves the lowest PCA compute time (red), owing to non-overlapping patches, but its training time surpasses that of Global PCA due to the need to train both the base PCA-Net and the refinement model. Despite this elevated training component, the total runtime remains the smallest among all approaches.

The figure also annotates each bar with performance metrics and parameter counts. Interestingly, L2L + Overlap attains marginally lower MSE and MAE values than Global PCA, illustrating that a slight increase in PCA computation and total parameters can yield improved reconstruction fidelity. Meanwhile, L2L + RefinementNet offers the fastest end-to-end process at the cost of slightly higher MSE and MAE. Thus, depending on whether the primary objective is to minimize total runtime or maximize solution accuracy, this figure highlights the trade-offs inherent to each approach. 

To benchmark our PCA-based patchwise approaches, we compare them against the state-of-the-art Multi-Grid Tensorized Fourier Neural Operator (MG-TFNO) framework \cite{kossaifi2023multigridtensorizedfourierneural}. MG-TFNO combines the expressive power of Fourier Neural Operators with a multi-resolution tensor factorization scheme (Tucker decomposition), allowing for efficient learning across multiple spatial scales while significantly reducing the number of trainable parameters.

We evaluate MG-TFNO models with three different tensor rank settings (0.5, 0.25, and 0.1), and compare them against our PCA-based patching approaches: one with overlapping patches and the other with a refinement network built on top of non-overlapping PCA patches. For all methods, the fields were divided into patches of size $32 \times 32$, and the training setup was kept consistent: a batch size of 32, initial learning rate of $1 \times 10^{-3}$ with ReduceLROnPlateau scheduler (patience = 30), L2 regularization of $1 \times 10^{-4}$, and training for 500 epochs.

\begin{figure}
    \centering
    \includegraphics[width=0.75\linewidth]{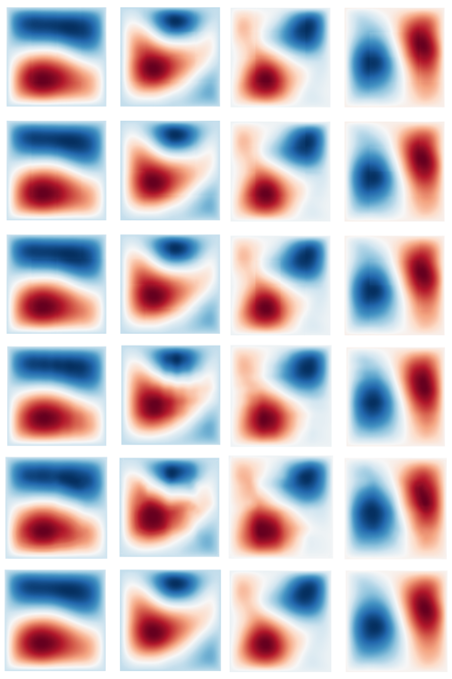}
    \caption{Comparison of reconstruction quality across MG-TFNO and Patch-PCA models. 
        Rows 1–3 show reconstructions obtained from the Multi-Grid Tensorized Fourier Neural Operator (MG-TFNO) with tensor ranks 0.1 (Row 1), 0.25 (Row 2), and 0.5 (Row 3), respectively. 
        Row 4 presents reconstructions from the Patch-PCA Overlap approach with an overlap size of 16. 
        Row 5 depicts results from the CNN-based RefinementNet with a convolutional kernel size of 5. 
        The final row (Row 6) displays the ground truth solution fields.}
    \label{fig:TFNO_comp}
\end{figure}

Sample reconstructions are illustrated in Fig. \ref{fig:TFNO_comp}: rows 1–3 show reconstructions from MG-TFNO with ranks 0.1, 0.25, and 0.5 respectively; row 4 corresponds to the patch PCA method with overlap; row 5 shows results from the refinement network; and the final row displays the ground truth fields. The first two columns highlight cases where MG-TFNO performs competitively, while the last two columns showcase examples where the PCA-based models achieve more accurate reconstructions.

A more quantitative comparison is presented in Table \ref{tab:appr_comp_table}. While MG-TFNO with higher ranks achieves strong SSIM values (up to 0.992), our PCA-based models demonstrate superior performance with significantly fewer trainable parameters and lower input feature dimensionality—most notably, the Patch-PCA with Overlap achieves an SSIM of 0.997 with only 114k parameters, and the RefinementNet variant attains 0.996 with just 144 input features per sample.

\begin{table}[h]
    \centering
    \small
    
        \begin{tabular}{lccc}
        \toprule
        \textbf{Model} & \textbf{Parameters} & \textbf{Input Features} & \textbf{SSIM} \\
        \midrule
        MG-TFNO (rank 0.5)             & 1.240M  & 6912 & 0.992 \\
        MG-TFNO (rank 0.25)            & 0.610M & 6912 & 0.986 \\
        MG-TFNO (rank 0.1)             & 0.290M & 6912 & 0.984 \\
        Patch-PCA w/ Overlap           & 0.114M & 487  & 0.997 \\
        Patch-PCA (No Overlap)       & 0.111M & 144  & 0.996 \\
        \bottomrule
        \end{tabular}
        \caption{Comparison of model sizes, input dimensionality (per sample), and SSIM performance across MG-TFNO and Patch-PCA variants.}
        \label{tab:appr_comp_table}
\end{table}

\section{Summary}

This work introduced a series of patch-based PCA-Net approaches aimed at learning efficient neural operator mappings for the two-dimensional Poisson equation. The primary challenge lies in mitigating the prohibitively high computational cost of performing PCA on large-scale, $128\times128$ solution fields. To address this issue, two fundamental patch-based configurations were investigated. In the Local-to-Global approach, the input fields were divided into smaller patches for individual PCA computations while the output solutions relied on a single, global PCA. This strategy effectively reduced PCA costs for the inputs and preserved global coherence in the solutions, albeit at the expense of the global PCA step remaining computationally intensive. In contrast, the Local-to-Local  configuration applied patch-based PCA on both inputs and outputs, drastically reducing PCA times across the board yet introducing blockiness artifacts in the reconstructed solution fields.

To mitigate the blockiness arising from the Local-to-Local method, two complementary refinements were proposed. First, introducing overlapping patches (smaller strides) and applying a Hanning window during reassembly produced smoother fields, often surpassing even the performance of global PCA on key metrics. Second, a RefinementNet approach was introduced, wherein a lightweight convolutional neural network was trained to smooth the initial, patch-based reconstruction, thereby reducing discontinuities and improving visual quality. Although this two-stage method required additional training time and slightly increased parameters, it led to faster overall processing when considering the markedly reduced PCA compute phase.

Extensive parametric studies evaluated the impact of patch sizes, stride lengths, and CNN kernel sizes. Moderate patch sizes (e.g., $16\times16$) typically yielded a favorable balance between dimensionality reduction efficiency and reconstruction accuracy. Meanwhile, among the kernel sizes tested, a $5\times5$ receptive field consistently delivered robust results. These experiments revealed that while larger patches minimize the total patch count, they increase individual PCA complexity, and while smaller patches or smaller strides boost reconstruction fidelity, they also require more patches and thus longer processing times.

Overall, this study demonstrates that patch-based PCA-Net frameworks substantially lessen the computational overhead of global PCA without compromising solution fidelity. The presence of local blockiness is addressed effectively by overlapping patches and smoothing, as well as by the RefinementNet’s secondary CNN refinement. Quantitatively, the end-to-end pipeline time—including PCA computation, training, and inference—was reduced by a factor of approximately 3.7–4× compared to the baseline global PCA pipeline. Practitioners can select or combine these refinements based on desired trade-offs between reconstruction accuracy, runtime, and training complexity. Future work could explore adaptive patch sizing, incorporate physics-informed loss functions, and extend these techniques to more complex PDE systems and higher-resolution datasets.

\section*{Acknowledgements}
This work was supported in part by the AFOSR Grant FA9550-24-1-0327.

\bibliographystyle{unsrt}
\bibliography{references}

\end{document}